\documentclass[conference]{IEEEtran}
\IEEEoverridecommandlockouts
\usepackage{cite}
\usepackage{amsmath,amssymb,amsfonts}
\usepackage{algorithmic}
\usepackage{graphicx}
\usepackage{subfig}       
\usepackage{textcomp}
\usepackage{xcolor}
\usepackage{adjustbox}
\usepackage{float}
\usepackage{multirow}

\def\BibTeX{{\rm B\kern-.05em{\sc i\kern-.025em b}\kern-.08em
    T\kern-.1667em\lower.7ex\hbox{E}\kern-.125emX}}
\begin{document}

\newcommand{\confName}{ICCIT }

\title{Maximizing Generalization: The Effect of Different Augmentation Techniques on Lightweight Vision Transformer for Bengali Character Classification\\

}


\author{ 
    \IEEEauthorblockN{
    Rafi Hassan Chowdhury\textsuperscript{1}, 
    Naimul Haque\textsuperscript{1}, 
    Kaniz Fatiha\textsuperscript{1}\\}

    \IEEEauthorblockA{
        \textsuperscript{1}Department of Computer Science and Engineering, Islamic University of Technology, Gazipur 1704, Bangladesh}

    \IEEEauthorblockA{\{rafihassan, naimulhaque, kanizfatiha\}@iut-dhaka.edu
    } 
}


\makeatletter
\let\old@ps@IEEEtitlepagestyle\ps@IEEEtitlepagestyle
\def\confheader#1{%
    \def\ps@IEEEtitlepagestyle{%
        \old@ps@IEEEtitlepagestyle%
        \def\@oddhead{\strut\hfill#1\hfill\strut}%
        \def\@evenhead{\strut\hfill#1\hfill\strut}%
    }%
    \ps@headings%
}
\makeatother
\confheader{
        \parbox{20cm}{2024 27th International Conference on Computer and Information Technology (ICCIT)\\
        20-22 December 2024, Cox’s Bazar, Bangladesh}
}

\IEEEpubid{
\begin{minipage}[t]{\textwidth}\ \\[10pt]
      \small{979-8-3315-1909-4/24/\$31.00 \copyright2024 IEEE }
\end{minipage}
}

\maketitle

\begin{abstract}
Deep Learning Models have proven to be highly effective in
computer vision, with deep convolutional neural networks achieving impressive results on various Computer Vision tasks.
However, they rely strongly on having large datasets to avoid
overfitting. When a model learns features with either low or high variance, it leads underfitting or overfitting on the training data. Unfortunately, wide-ranging dataset might not exist in many fields such as resource limited languages like Bengali. In our experiment, a series of the tests were conducted in the field of Image data augmentation as
an approach to addressing the limited data problem for Bengali handwritten characters. It also provides an in-depth analysis of the performance of these augmentation techniques. Data augmentation refers to a set of techniques or procedures applied to data to make them
larger in size and more precise, which make them more suitable for
training deep learning models. The image augmentation techniques evaluated in this study include CLAHE, Random Rotation,
Random Affine, Color Jitter, and thier combinations. The study also
explores the use of augmentation methods in lightweight model
like EfficientViT. Among different augmentation methods,
we found combination of Random Affine and Color Jitter produced the best accuracy in Ekush\cite{ekush} and AIBangla\cite{9084029} datasets, with accuracy of 97.48\% and  97.57\% respectively.
This combination outperformed all other combinations, with or without augmentation techniques. This analysis presents a thorough examination of the impact of image data augmentation in resource-scarce languages, particularly in the context of Bengali handwritten characters recognition using lightweight model.

\end{abstract}


\section{Introduction}
Deep learning models have demonstrated exceptional performance in discriminative tasks, largely due to advancements in deep architectures and the substantial computational resources available, which are further enhanced by the vast quantities of big data. CNNs have been central to the story of how AI has advanced for many fields with a strong emphasis on Visual tasks like image classification, object detection, and segmentation. CNNs are also able to maintain spatial properties of images by using tightly parameterized but sparsely connected kernels. Convolutional layers are used to extract features, they take advantage of the local spatial coherence and reduces the dimension of features maps while increasing their depth, hence provide a localized version of input image that allows meaningful representations rather than hand engineered features. The groundbreaking success of CNNs has lead to numerous efforts in using Deep Learning techniques for various Computer Vision problems.

Since it is difficult to enhance the generalization capability of those models. In simple words, Generalization refers to how well the model performs on unseen data, relative to its performance on training set. Sometimes models become over-fitted. When a model performs good on the training data but performs poor when it comes to predict or classify unseen data, this term is known as overfitting. One way is utilize early stopping use validation set to overcome overfitting while training our models.

A Deep Learning model is learning effectively when both training error and validation error are decreasing. One of the appropriate ways to maintain proper training, is by increasing the dataset size using data augmentation techniques. It is important that augmented data are not too different from existing data, so that the learning process is generalize. This allows the model to cover various cases possible with a large dataset, which helps to form larger training and validation sets that is immediately adjacent in distribution. In future, the testing set also gets benefit from this approach.

Bangla is the seventh most spoken language in the world, but despite of its widespread usage, if faces unique challenges due to the lack of diverse and high-volume datasets. Existing datasets are narrow in scope and size. Creation of a large-scale and high quality dataset would be ideal but it is actually quite costly and time consuming to annotate. On the other hand, image data augmentation provides an economical alternative to achieve improved model performances, and has proven that augmentation improved model performance \cite{Nagaraju2022}. Artificially generating data samples that follow the representation of the original set helps maintain nuances in Bangla characters. Due to the lack of image data for the Bangla language, this type of augmentation technique allows models to generalize better and become more robust. While there have been some attempts at image augmentation for Bengali datasets, a detailed evaluation of which augmentation methods best preserve the integrity of this resource-constrained language is still lacking.

We have covered related studies in Section \ref{Literature Review}. In Section \ref{Methodology} we discussed the different components of our experiment, and in Section \ref{Results and Discussion} the results are presented. At the end, we concluded the work in Section \ref{conclusion} by describing potential future directions.

\section{Literature Review}\label{Literature Review}
In recent years, the use of data augmentation methods has proven effective in boosting model generalization and reducing computational requirements, particularly for Bengali word recognition tasks. Howard et al. \cite{howard2017mobilenets} pioneered MobileNets, a set of lightweight neural network models that are ideal for mobile and embedded systems. Their work shows that small scale architectures, when enhanced with variations in data such as random cropping and image flipping can perform on par with more complex models.

Szegedy et al. \cite{7780677} built upon these foundations by introducing the Inception network, which incorporates a variety of advanced augmentation techniques. These include multi-scale cropping and color distortions, which significantly enhance the generalization capacity of the network. Their research is crucial for understanding how augmentations help neural networks, including compact ones, resist overfitting. In another study, \cite{perez2017effectivenessdataaugmentationimage} the authors evaluate various approaches to address the challenge of data augmentation in image classification where they intentionally limit their data access to a small portion of the ImageNet dataset and sequentially assess each data augmentation method and also they explore the use of Generative Adversarial Networks (GANs) to create images in different styles.

Focusing on Bengali word recognition, Chowdhury et al. \cite{8858545} investigated the effects of geometric transformations and changes in contrast, finding that these augmentation techniques improve the accuracy of handwritten character recognition by minimizing overfitting and this highlights the importance of customized augmentation strategies, especially in cases where data collection is limited, as it often is for under-resourced languages. In another work, \cite{RAQUIB2024100568} they introduced an automated system for recognizing handwritten Bangla basic characters called VashaNet. They use deep convolutional neural network.

In another study, Lancheros et al. \cite{app11156845} explored augmentations in the biomedical field utilizing Transfer Learning for Cross-Lingual Named Entity Recognition where they introduced a new augmented dataset is created by altering 20\% of the entities in the original dataset. 


In another study, BengaliNet an innovative, cost-effective convolutional neural network architecture designed for the recognition of Bengali characters, utilizing a minimal number of parameters relative to the number of output classes where in the training phase, they examined eight distinct dataset configurations derived from prior research. \cite{app11156845}

\section{Methodology}\label{Methodology}

For our experiment, we initially split the dataset into training, validation, and test sets with a ratio of 60:20:20. Subsequently, we created every possible combination of our augmentation techniques. We then trained the EfficientViT model using pretrained weights for each combination of augmentation techniques. Finally, we analyzed the results, which are summarized in Table \ref{fig:ekush_dataset_sample} and Table \ref{tab:augmentation_aibangla}.

\subsection{Datasets}
\begin{figure}[htbp] 
    \centering
    \includegraphics[width=0.28\textwidth]{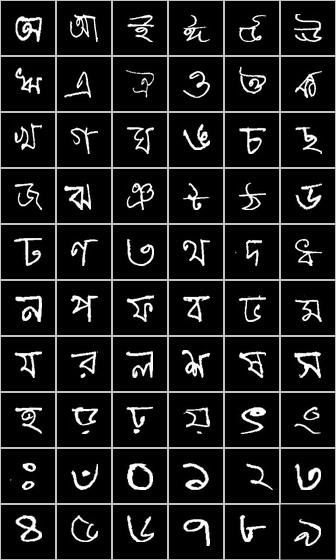}
    \caption{Handwritten Bangla Basic Characters and Numerals from Ekush Dataset}
    \label{fig:ekush_dataset_sample}
\end{figure}

\begin{figure}[htbp] 
    \centering
    \includegraphics[width=0.28\textwidth]{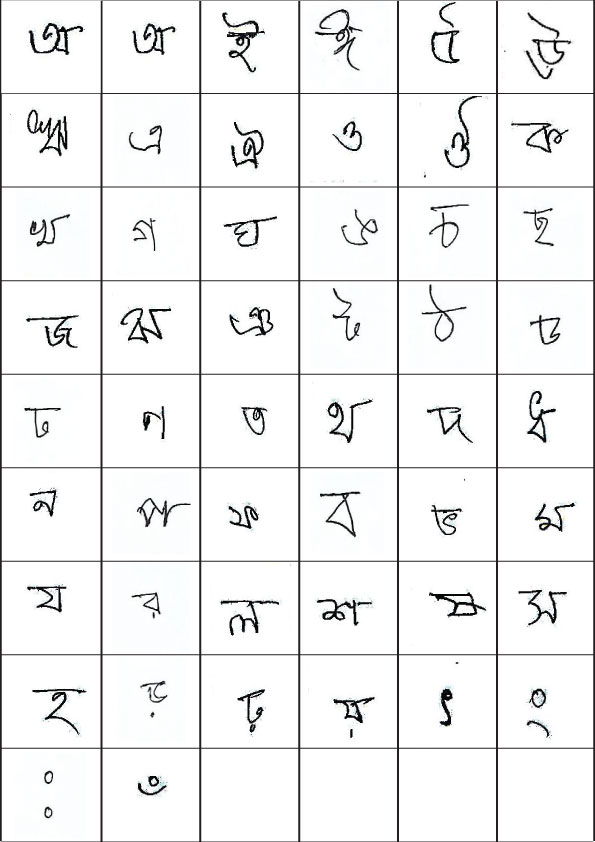}
    \caption{Handwritten Bangla Basic Characters from AI-bangla Dataset}
    \label{fig:aibangla_dataset_sample}
\end{figure}

Quite a few open-access comprehensive datasets are available for handwritten Bangla character recognition, each addressing various challenges in the domain.  Ekush \cite{ekush}, and AIBangla \cite{9084029} are mentionable among others. Ekush was introduced by Rabby et al. \cite{ekush}, featuring 367,018 images, making it the largest dataset for handwritten Bangla characters. The dataset includes modifiers, vowels, consonants, compound characters, and numerals, written by 3,086 unique individuals across diverse age groups, genders, and districts within Bangladesh. Ekush focuses on character recognition and also supports gender and age prediction based on handwriting .  It is used for robust training and validation of deep learning models in Bangla OCR and related tasks. AIBangla, created by Hasan et al. \cite{9084029}, contributes significantly by providing over 330,000 images of handwritten Bangla characters, covering both basic and compound characters. It provides a large enough sample size to train complex models for accurate character recognition. AIBangla, created by 2000 contributors, has diverse writing styles that aid in researching handwritten Bangla character recognition using modern CNNs . A comparison between the three datasets are given in Table. 1. For our experiment, we have used Basic Characters from both datasets. Figure \ref{fig:aibangla_dataset_sample} and Figure \ref{fig:ekush_dataset_sample} showing samples from AIBangla and Ekush datasets respectively. 

\begin{table}[htbp]
\caption{Comparison of Bangla Handwritten Character Datasets}
\centering
\begin{adjustbox}{width=\linewidth}
\begin{tabular}{|l|c|c|c|}
\hline
\textbf{Dataset Name}       & \textbf{Modifiers} & \textbf{Basic Characters} & \textbf{Compound Characters} \\ \hline
\textbf{Ekush}               & 30,667            & 154,824                   & 150,840                      \\ 
\textbf{AIBangla Basic}            & None              & 80,403                    & 249,911                      \\ \hline
\end{tabular}
\end{adjustbox}

\label{tab:datasets}
\end{table}

\subsection{Model Architecture}

For this study we use light weight model. We chose EfficientViT as our lightweight Vision Transformer model because it strikes an excellent balance between efficiency and performance. While MobileViT and TinyViT are both designed to be compact, EfficientViT stands out with its clever architectural innovations that improve computational efficiency without sacrificing accuracy. This makes it particularly well-suited for deployment in environments with limited resources. Furthermore, EfficientViT is scalable and adaptable, performing well across various vision tasks, which allows us to easily integrate it into different applications. With its ability to deliver low latency and effectively utilize parameters while maintaining strong performance, EfficientViT emerged as the best fit for our lightweight vision needs.

EfficientViT is designed for high-resolution dense prediction tasks with an efficient and scalable architecture where the central innovation is the multi-scale linear attention mechanism that Han et al.\cite{cai2022efficientvit} apply in order to effectively trade off cost of computation and performance as well. The model contains 2.14 million trainable parameters, has a size of 2.04 MB, and operates at 0.1 GFLOPs, making it lightweight and efficient. Traditional softmax-based attention mechanisms usually have quadratic complexity w.r.t input resolution, which makes them not applicable for high-resolution tasks EfficientViT, on the other hand, leverages the associative property of matrix multiplication by using ReLU-based linear attention, turning its complexity to be only quadratic. This change enables EfficientViT to run on high resolution images without experiencing the compute bottleneck expected by others models. The multi-scale linear attention aggregates nearby tokens using small-kernel convolutions to generate multi-scale tokens, effectively capturing both local and global features. This multi-scale approach is crucial for dense prediction tasks, where understanding fine details and broader contexts is essential. In super-resolution tasks, the enforcement provides a 8.8× GPU latency reduction over SegFormer and 6.4× speedup in excess of Restormer, allowing it even more ideal regarding real-time purposes. This efficiency, coupled with its ability to capture both global and local contexts, positions EfficientViT as a state-of-the-art solution for high-resolution dense prediction tasks.

The architecture of EfficientViT is elegantly crafted around the EfficientViT Module, following a backbone-head/encoder-decoder design that is prevalent in modern vision transformer models, the backbone begins with an input stem and is organized into four stages, each progressively reducing the feature map size while simultaneously increasing the channel count and this hierarchical design allows for efficient feature extraction and representation.\cite{cai2022efficientvit} Notably, the EfficientViT Module is integrated into Stages 3 and 4, enhancing the model's capability to capture intricate patterns within the data. To facilitate downsampling, we employ an MBConv with a stride of 2, ensuring that features are effectively distilled at each stage. The head of the architecture synthesizes outputs from Stages 2, 3, and 4 designated as P2, P3, and P4 creating a pyramid of feature maps. For added efficiency and simplicity, we utilize 1x1 convolutions in conjunction with standard upsampling techniques, such as bilinear or bicubic upsampling. This thoughtful architecture not only empowers EfficientViT to adeptly process high-resolution inputs but also fosters effective feature aggregation, positioning it as a robust solution for dense prediction tasks.

\subsection{Augmentation Methods}

\subsubsection{CLAHE} Contrast Limited Adaptive Histogram Equalization (CLAHE) by Zuiderveld \cite{zuiderveld1994contrast} is an extension of adaptive histogram equalization to prevent the over-amplification of noise in images. It increases local contrast, particularly in images with varying brightness levels. It divides the image into small tiles and applies HE separately to each tile, effectively limiting the contrast amplification\cite{xu2023comprehensive}. The limitation is achieved by clipping the histogram at a specified threshold. CLAHE can be represented as: 
\[
G(i,j) = \frac{L-1}{MN} \sum_{k=0}^{i} \sum_{i=0}^{j} p(f(k,l),C)
\]

CLAHE has been applied in CNN-based architectures to enhance image data for training deep learning models, especially in cases where data is limited or imbalanced. Here Figure \ref{fig:aug_sample} (e) is a sample of CLAHE. 

\subsubsection{Random Rotation}
Rotation augmentations involve rotating the image right or left on an axis by an angle between 1° and 359° \cite{kamra2024simsamsimplesiameserepresentations}. Random Rotation is a technique that involves applying random rotations to images within a specified range of degrees, typically to introduce variety and diversity in the dataset during training. This helps the model become more robust and generalize better to unseen data. It is most useful in situations where objects in the real world are oriented differently, but the data collected is limited to a fixed perspective \cite{kamra2024simsamsimplesiameserepresentations}. By simulating these variations, random rotation helps the model perform better without needing additional data collection and labeling. Studies, such as those by Muñoz-Aseguinolaza et al. \cite{kamra2024simsamsimplesiameserepresentations}, demonstrate how this technique helps reduce overfitting and improves performance in image recognition tasks. In our experiment, we have used random rotation in the range of -45 degrees to +45 degrees. Figure \ref{fig:aug_sample} (b) is a sample of random rotation.

\subsubsection{Random Affine}
Random affine includes random geometrical operations such as rotation, translation, scaling, and shearing, which keep the relative locations of the points of the image intact \cite{kamra2024simsamsimplesiameserepresentations}. The randomness of these processes introduces diversity in the training set that will enhance generalization and reduce overfitting in the models. \cite{kamra2024simsamsimplesiameserepresentations}. In our experiment, we used height and width shifts in the range of 0 to 0.1 and applied shear with a factor of 20. A sample of Width-Height and Shearing is showing in Figure \ref{fig:aug_sample} (c) and Figure \ref{fig:aug_sample} (d) respectively.Random affine transformations are remarkably effective in applications with underlying data possessing inherent spatial variability. Affine transformation fortifies a model by making it robust against changes in patient placement or imaging conditions in medical imaging \cite{shao2023augdiffdiffusionbasedfeature}. The complexity of Bengali characters, with their ligatures and diacritical marks, leads to significant variability in handwriting. Techniques like random affine transformations—such as rotation, scaling, and translation—are applied during neural network training, particularly in CNNs, to account for these variations. This allows the model to recognize characters even when distorted by diverse handwriting styles.

\subsubsection{Color Jitter} 
\begin{figure}[H]
    \centering
    \begin{tabular}{cccccc}
        \subfloat[Original]{\includegraphics[width=0.15\textwidth]{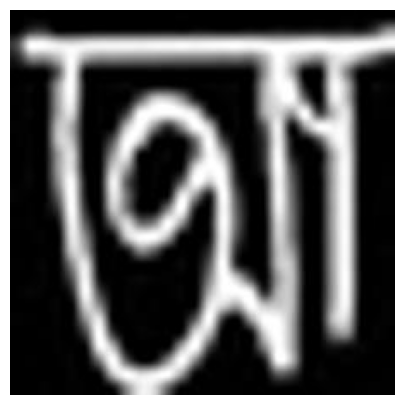}} &
        \hspace{-0.40cm} 
        \subfloat[Random Rotation]{\includegraphics[width=0.15\textwidth]{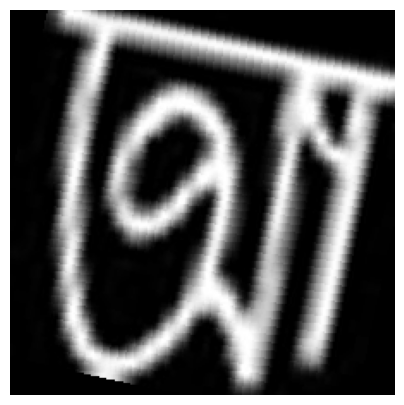}} &
        \hspace{-0.40cm} 
        \subfloat[Width \& Height shift]{\includegraphics[width=0.15\textwidth]{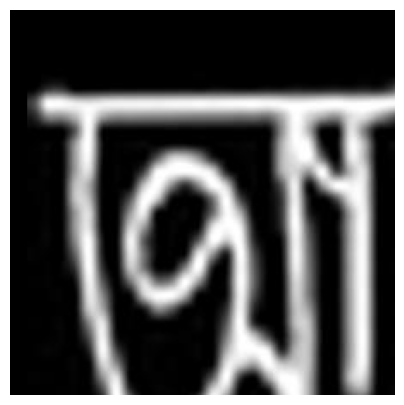}} &
        \hspace{-0.40cm} 
        \\
        \subfloat[Shearing]{\includegraphics[width=0.15\textwidth]{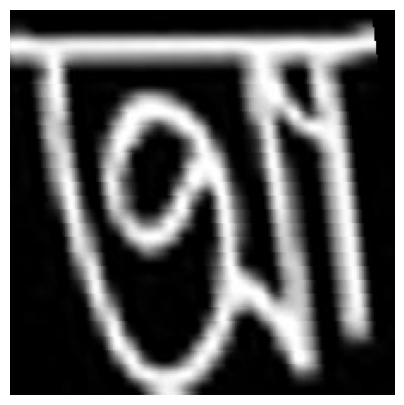}} &
        \hspace{-0.40cm} 
        \subfloat[CLAHE]{\includegraphics[width=0.15\textwidth]{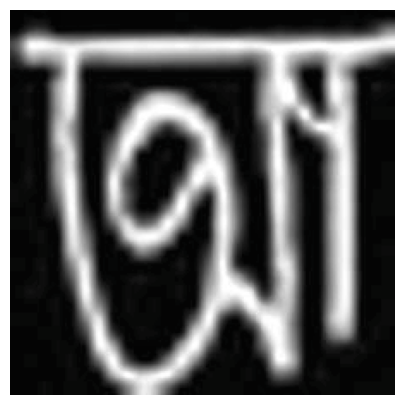}} &
        \hspace{-0.40cm} 
        \subfloat[Color Jitter]{\includegraphics[width=0.15\textwidth]{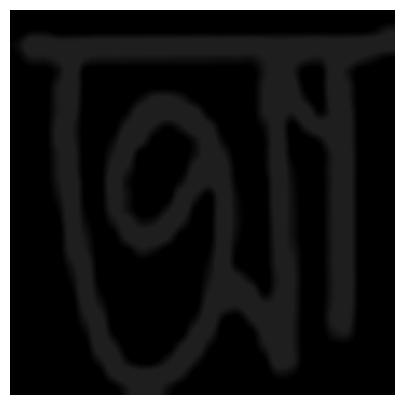}} \\
    \end{tabular}
    \caption{Sample of augmentation techniques}
    \label{fig:aug_sample}
\end{figure}

ColorJitter adds variations in RGB images to simulate diverse lighting conditions or color distortions \cite{inproceedings1}. This stochastic transformation approximates real-world changes a model may see in deployment and is, therefore, quite useful for enhancing the generalization of CNNs \cite{inproceedings2}. ColorJitter supports random selection of a factor within a given range by allowing setting of parameters. For instance, the brightness factor is chosen from [max(0, 1 - brightness), 1 + brightness] so that the image correctly darkens/lightens up. In order not to drastically change the colour, the value of hue parameter has to be in [-0.5, 0.5] and hence its value should be dealt with carefully.\cite{zini2023planckianjittercounteringcolorcrippling}. In our experiment, we have applied color jitter with configuration allowing brightness, contrast, and saturation to vary by ±20\%, and hue to change by ±10\%. Figure \ref{fig:aug_sample} (f) is a sample of color jitter.
ColorJitter enhances the resilience of CNNs to color distortions by introducing random color changes and helps them learn more robust features \cite{inproceedings2}. ColorJitter can add variation to the appearance of Bengali handwritten characters in training data, improving recognition under various lighting conditions. This is crucial because the complex curves of Bengali letters often lead to errors. By adjusting brightness, contrast, and saturation, the model becomes more resilient to differences in paper, ink, or scan quality.

\subsection{Experimental Setup}
Experiments were done on desktop with CPU as Core i5 13500 of 2.50 GHz and Nvidia RTX 3060 having 12 GB VRAM. Memory size was 32GB. The EfficientViT model is also trained on the Python environment. Training was performed using these Pre-trained Pytorch models from timm (ImageNet-1k pretrained) Cross Entropy loss was employed since it is suitable for multi class classification.

The models were trained for 100 epochs. By introducing cross-validation, we set early stopping to fire during training after an epoch if the current average validation loss was greater than that from any previous epoch. Training is stopped if five patients in a row are triggered as this was added to avoid overfitting of the models. They are generally converged within the 50 to 60 epochs for most of tests and models. Learning rate and Batch Size were 0.00001 and 64 respectively.

\section{Result Analysis} \label{Results and Discussion}

\subsection{Performance of our Method}
To evaluate the performance of our model, we have used the common metrics in deep learning like accuracy, precision, recall and F1-score for evaluating our model. These metrics were also used to provide a comprehensive evaluation framework of the proposed augmentation techniques over the various datasets used in this work.

\begin{table}[htbp]
\caption{Comparison of Augmentation Techniques for AIBangla\_Basic Dataset using EfficientViT}
\centering
\begin{tabular}{|l|c|c|c|c|}
    \hline
    \multirow{2}{*}{\textbf{Augmentation Technique}} & \multicolumn{4}{c|}{\textbf{AIBangla\_Basic}} \\
    \cline{2-5}
    & \textbf{precision} & \textbf{recall} & \textbf{f1-score} & \textbf{accuracy} \\
    \hline
    None                & 96\% & 96\% & 96\% & 96.36\% \\
    RandomRotation (RR) & 97\% & 97\% & 97\% & 96.92\% \\
    RandomAffine (RA)   & 97\% & 97\% & 97\% & 97.28\% \\
    CLAHE (C)           & 97\% & 97\% & 97\% & 96.51\% \\
    ColorJitter (CJ)    & 97\% & 97\% & 97\% & 96.65\% \\
    RR + RA             & 97\% & 97\% & 97\% & 96.67\% \\
    RR + C              & 97\% & 97\% & 97\% & 96.89\% \\
    RR + CJ             & 97\% & 97\% & 97\% & 96.75\% \\
    RA + C              & 98\% & 98\% & 98\% & 97.55\% \\
    RA + CJ             & 98\% & 98\% & 98\% & \textbf{97.57\%} \\
    C + CJ              & 97\% & 97\% & 97\% & 96.66\% \\
    RR + RA + C         & 97\% & 97\% & 97\% & 96.56\% \\
    RR + RA + CJ        & 96\% & 96\% & 96\% & 96.26\% \\
    RR + C + CJ         & 97\% & 97\% & 97\% & 96.83\% \\
    RA + C + CJ         & 97\% & 97\% & 97\% & 97.42\% \\
    RR + RA + C + CJ    & 97\% & 97\% & 97\% & 96.71\% \\
    \hline
\end{tabular}
\label{tab:augmentation_aibangla}
\end{table}

\begin{table}[htbp]
\caption{Comparison of Augmentation Techniques for Ekush-Bangla Dataset using EfficientViT}
\centering
\begin{tabular}{|l|c|c|c|c|}
    \hline
    \multirow{2}{*}{\textbf{Augmentation Technique}} & \multicolumn{4}{c|}{\textbf{Ekush-Bangla}} \\
    \cline{2-5}
    & \textbf{precision} & \textbf{recall} & \textbf{f1-score} & \textbf{accuracy} \\
    \hline
    None                & 97\% & 97\% & 97\% & 97.23\% \\
    RandomRotation (RR) & 97\% & 97\% & 97\% & 97.22\% \\
    RandomAffine (RA)   & 97\% & 97\% & 97\% & 97.45\% \\
    CLAHE (C)           & 97\% & 97\% & 97\% & 97.24\% \\
    ColorJitter (CJ)    & 97\% & 97\% & 97\% & 97.34\% \\
    RR + RA             & 97\% & 97\% & 97\% & 96.99\% \\
    RR + C              & 97\% & 97\% & 97\% & 97.16\% \\
    RR + CJ             & 97\% & 97\% & 97\% & 97.12\% \\
    RA + C              & 98\% & 98\% & 98\% & 97.40\% \\
    RA + CJ             & 98\% & 97\% & 97\% & \textbf{97.48\%} \\
    C + CJ              & 97\% & 97\% & 97\% & 97.28\% \\
    RR + RA + C         & 97\% & 97\% & 97\% & 96.89\% \\
    RR + RA + CJ        & 97\% & 97\% & 97\% & 97.05\% \\
    RR + C + CJ         & 97\% & 97\% & 97\% & 97.33\% \\
    RA + C + CJ         & 97\% & 97\% & 97\% & 97.40\% \\
    RR + RA + C + CJ    & 97\% & 97\% & 97\% & 97.06\% \\
    \hline
\end{tabular}
\label{tab:augmentation_ekushbangla}
\end{table}

In our experiment, four image augment techniques i.e. Random Rotation (RR), Random Affine (RA), CLAHE (C), and ColorJitter (CJ) are used along with multiple combinations of those techniques. Out of these, we can observe that the RA + CJ combination performed better compared to all other combinations on both datasets and also surpassed the reported best accuracy in previous studies with an 97.57\% accuracy score for AIBangla\_Basic and 97.48\% for Ekush-Bangla.

We get effective performance from RA + CJ, because the combination of these two augmentation techniques generating real time data and they are not changing the actual feature of the data. As a result our model is able to learn salient features from the data. However, other augmentation combinations reveal mixed results. RA alone performs admirably, reaching 97.28\% accuracy on AIBangla\_Basic and 97.45\% on Ekush-Bangla, illustrating the strength of geometric transformations in capturing salient feature in a character structure. 

RA and CJ work together and RA will be helpful in handling slight distortions while ColorJitter is capacitor enough to handle different ranges of lighting condition and minor color adjustment making model more robust in real time scenarios. Augmentation was not always the answer is our observations as well. However, if there is too much change from one type to the other then these could distort the input too much for a model to classify accurately. For example, Random Rotation when used together with other affine transformations can resemble too different characters for the model to understand. However, some methods like CLAHE which are useful to increase contrast may result in a loss of diversity for the augmented data together with other generated samples making the model to learn very few general features.

We conducted experiments utilizing three different ViT models, as shown in Table \ref{tab:comparison_of_performance_all_vit}, where EfficientViT outperformed the other models.

\begin{table}[htbp]
\caption{Performance comparison with Baseline Vision Transformer models of RA + CJ combination on Ekush-Bangla Datasets\cite{ekush} and AIBangla\_Basic\cite{9084029}.}
\centering
\begin{adjustbox}{width=\linewidth}
\begin{tabular}{l p{ 1.0cm} p{1.0cm} p{1.5cm} p{1.0cm} p{1.0cm}}
\hline
\textbf{Architecture} & \textbf{Accuracy \cite{ekush} (\%)}& \textbf{Accuracy \cite{9084029} (\%)} & \textbf{Trainable Parameters Count (M)} & \textbf{Model Size (MB)} & \textbf{FLOPs Count (GFLOPS)} \\ \hline

MobileViT   & 96.47 & 96.67  & 4.94  & 4.71 & 1.44 \\ 
TinyViT   & 97.39 & 97.07 & 5.07  & 4.84   & 1.17     \\
EfficientViT  &  \textbf{97.48} & \textbf{97.57} & \textbf{2.14}  & \textbf{2.04}  & \textbf{0.1}  \\ \hline

\end{tabular}
\end{adjustbox}
\label{tab:comparison_of_performance_all_vit}
\end{table}

\subsection{Comparison with state-of-the-art methods}

We conducted an experiment to demonstrate how data augmentation can impact the classification of Bangla handwritten characters. Very limited research has been conducted on this topic in previous for Bangla characters classification. In Table \ref{tab:comparison}, we show a comparision of our results using Ekush dataset. Our method achieved an accuracy of 97.48\%, while the DConvAENNet and CNN models only able to obtained accuracies of 95.53\% and 95.49\%, respectively.

\begin{table}[H]
\centering
\caption{Comparison with state-of-the-art methods on Ekush Dataset}
\label{table:performance_analysis_with_sota_models_Ekush}
\begin{tabular}{p{5.0cm} l }
\hline
\textbf{Architecture} & \textbf{Accuracy (\%)}  \\ \hline 
DConvAENNet \cite{azad2020bangla} & 95.53\% \\ 
CNN \cite{opu2024handwritten} & 95.49\% \\

\textbf{Ours} & \textbf{97.48\%} \\ \hline
\end{tabular}
\label{tab:comparison}
\end{table}

\subsection{Qualitative Analysis}

\begin{figure}[htbp]
    \centering
    \subfloat[Actual: \textit{chandra bindu}]{%
        \includegraphics[width=0.45\linewidth]{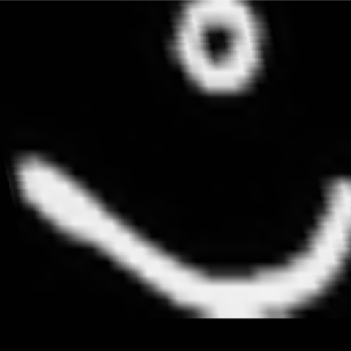}
        \label{fig:image1}
    }\hfill
    \subfloat[Predicted: \textit{chandra bindu}]{%
        \includegraphics[width=0.45\linewidth]{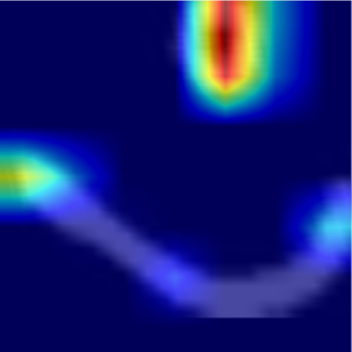}
        \label{fig:image2}
    }\\
    \subfloat[Actual: \textit{a}]{%
        \includegraphics[width=0.45\linewidth]{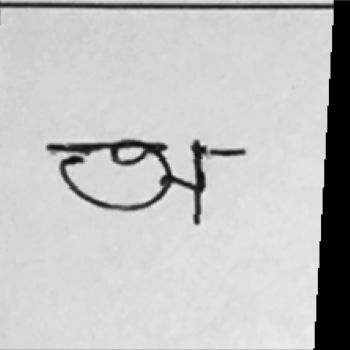}
        \label{fig:image3}
    }\hfill
    \subfloat[Predicted: \textit{a}]{%
        \includegraphics[width=0.45\linewidth]{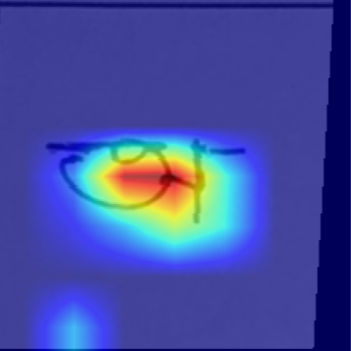}
        \label{fig:image4}
    }
    \caption{GradCam samples of correctly classified inputs}
    \label{fig:grad_cam}
\end{figure}

To understand if our proposed method is effectively identifying diseases using the important features, we use a technique called GradCAM. This method illustrates gradients of final layer, which helps us to visualize how well our model focuses on specific features of a character for correct classification, as shown in Figure \ref{fig:grad_cam}. This figure shows that each Bangla character has its own distinct pattern, and our model is capable of recognizing these patterns to do accurate classification.

\begin{figure}[htbp]
    \centering
    \subfloat[Predicted: \textit{dirgho\_i} Actual: \textit{rosho\_i}]{%
        \includegraphics[width=0.45\linewidth]{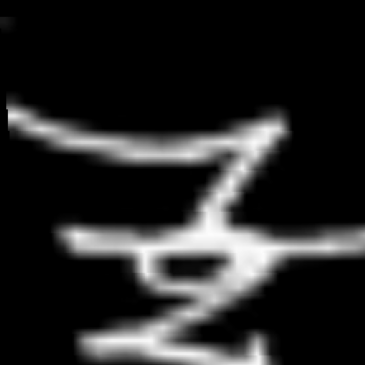}
        \label{fig:image1}
    }\hfill
    \subfloat[Predicted: \textit{jha} Actual: \textit{ri}]{%
        \includegraphics[width=0.45\linewidth]{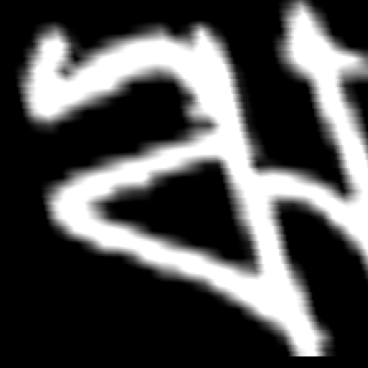}
        \label{fig:image2}
    }\\
    \subfloat[Predicted: \textit{e} Actual: \textit{ai}]{%
        \includegraphics[width=0.45\linewidth]{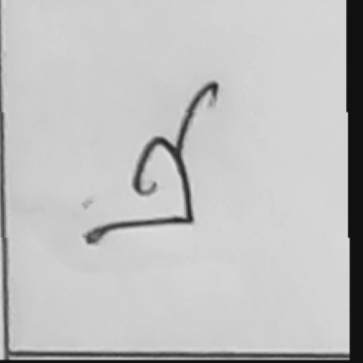}
        \label{fig:image3}
    }\hfill
    \subfloat[Predicted: \textit{ba} Actual: \textit{ka}]{%
        \includegraphics[width=0.45\linewidth]{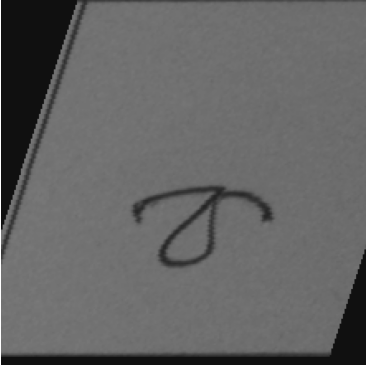}
        \label{fig:image4}
    }
    \caption{Misclassified samples that are visually similar to the predicted class.}
    \label{fig:missclassified_fig}
\end{figure}

After analyzing the misclassified samples, we found out tha int most of the cases, the reason for misclassification is inter-class similarity. Figure \ref{fig:missclassified_fig} illustrates some examples of inter-class similarity. In Figure \ref{fig:missclassified_fig} (c) we see \textbf{"ka"}, which is similar to \textbf{"ba"}. This is almost true for all misclassified samples, where the main reason in inter-class feature and pattern similarity.

\section{Conclusion \& Future work} \label{conclusion}

In our exploration of Bangla handwriting recognition, we conducted a study of the effectiveness of different augmentation techniques but it was important to start from lightweight model as it is more accessible for the developing countries which have resource limitations. In general, we observed an increase in model accuracy through the use of augmentation with Random Affine (RA) combined with Color Jitter (CJ) being one of the most effective augmentation on both datasets. 

In future, we can explore other augmentation techniques such as GAN, fliping, noise injection, cropping etc. for both lightweight and heavyweight models. Extending to more combinations of varied complexity, These future experiments will help not only to provide insight into effective augmentation strategies, but also aid in the construction of more generalize
models able to function well in complex and noisy environments.



\bibliographystyle{ieeetr}
\bibliography{citations} 

\end{document}